\documentclass[letterpaper]{article} 
\usepackage{aaai25}  
\usepackage{times}  
\usepackage{helvet}  
\usepackage{courier}  
\usepackage[hyphens]{url}  
\usepackage{graphicx} 
\usepackage{natbib}  
\usepackage{caption} 
\usepackage{algorithm}
\usepackage{algorithmic}

\usepackage{booktabs}
\usepackage{colortbl}
\usepackage{xcolor}
\usepackage{subfigure}
\usepackage{amsmath}
\usepackage{makecell}
\usepackage{multicol}
\usepackage{multirow}
\usepackage{amssymb}

\usepackage{newfloat}
\usepackage{listings}
\DeclareCaptionStyle{ruled}{labelfont=normalfont,labelsep=colon,strut=off} 
\floatstyle{ruled}
\newfloat{listing}{tb}{lst}{}
\floatname{listing}{Listing}
\title{Bridging Molecular Graphs and Large Language Models}

\author {
    Runze Wang\textsuperscript{\rm 1},
    Mingqi Yang\textsuperscript{\rm 2},
    Yanming Shen\textsuperscript{\rm 1}\thanks{Corresponding author.}
}
\affiliations {
    \textsuperscript{\rm 1}School of Computer Science and Technology, Dalian University of Technology, China\\
    \textsuperscript{\rm 2}School of Computing, National University of Singapore, Singapore\\
    runze\_wang@mail.dlut.edu.cn, 
    mqyang@nus.edu.sg, 
    shen@dlut.edu.cn
}

\usepackage{bibentry}

\begin{document}

\maketitle

\begin{abstract}
While Large Language Models (LLMs) have shown exceptional generalization capabilities, their ability to process graph data, such as molecular structures, remains limited.
To bridge this gap, this paper proposes Graph2Token, an efficient solution that aligns graph tokens to LLM tokens.
The key idea is to represent a graph token with the LLM token vocabulary, without fine-tuning the LLM backbone.
To achieve this goal, we first construct a molecule-text paired dataset from multi-sources, including CHEBI
and HMDB, to train a graph structure encoder, which reduces the distance between graphs and texts representations in the feature space. 
Then, we propose a novel alignment strategy that associates a graph token with LLM tokens. 
To further unleash the potential of LLMs, we collect molecular IUPAC name identifiers, which are incorporated into the LLM prompts. 
By aligning molecular graphs as special tokens, we can activate LLMs' generalization ability to molecular few-shot learning.
Extensive experiments on molecular classification and regression tasks demonstrate the effectiveness of our proposed Graph2Token.
\end{abstract}

%

\begin{links}
    \link{Code}{https://github.com/GraphMoLab/Graph2Token}
\end{links}

\section{Introduction}

Recent studies have shown promising results in applying large language models (LLMs) to graph machine learning, particularly demonstrating the potential in few-shot or zero-shot learning on knowledge graphs and text-attributed graphs \cite{fan2024graph, tang2024graphgpt, chenllaga}.
However, their capability to handle graph-structured data such as molecules is still not well investigated. 
The intrinsic properties of molecules demand a deeper understanding of their structures beyond node attributes, posing a challenge that requires a distinct approach compared to handling text attribute graphs.
Extending the functionality of LLMs to effectively process and analyze molecules will open up opportunities for molecular related tasks.

\begin{figure}[t]
\centering
\subfigure[SMILES as input]{
\includegraphics[width=3.8cm]{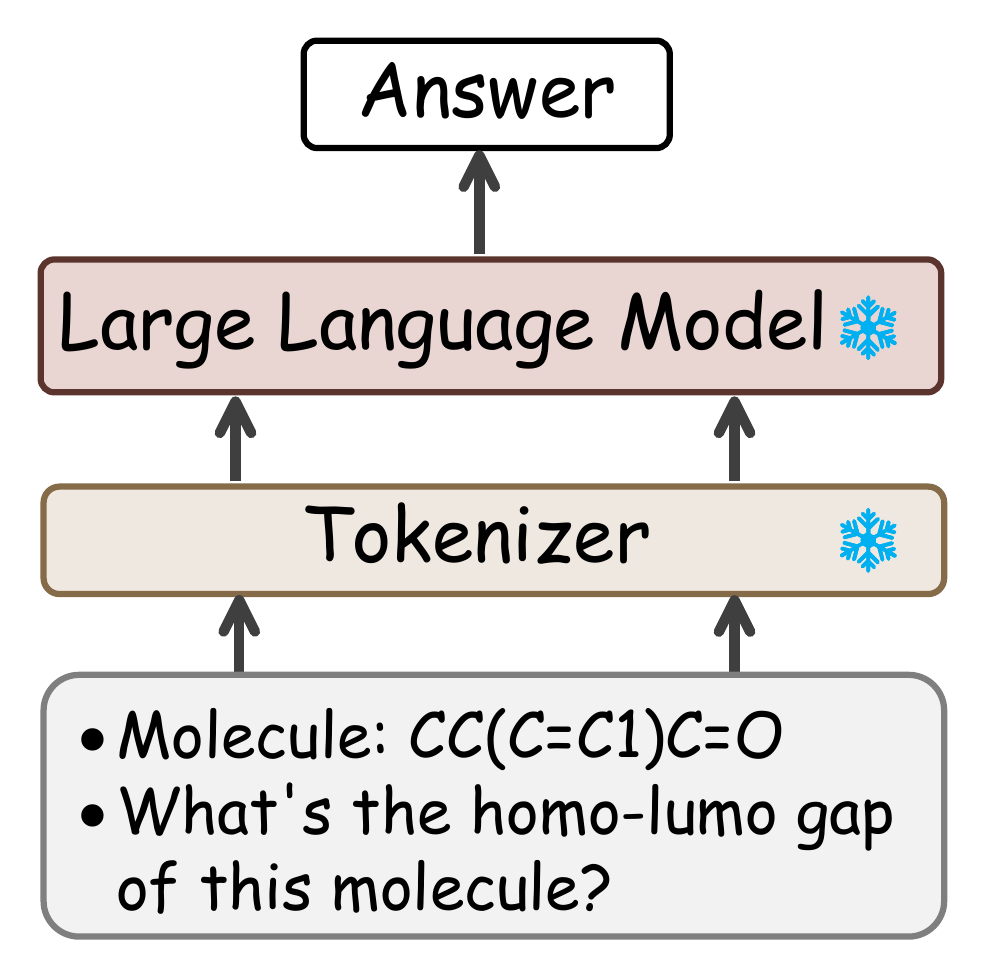}
}
\quad
\subfigure[Textualized graph as input]{
\includegraphics[width=3.8cm]{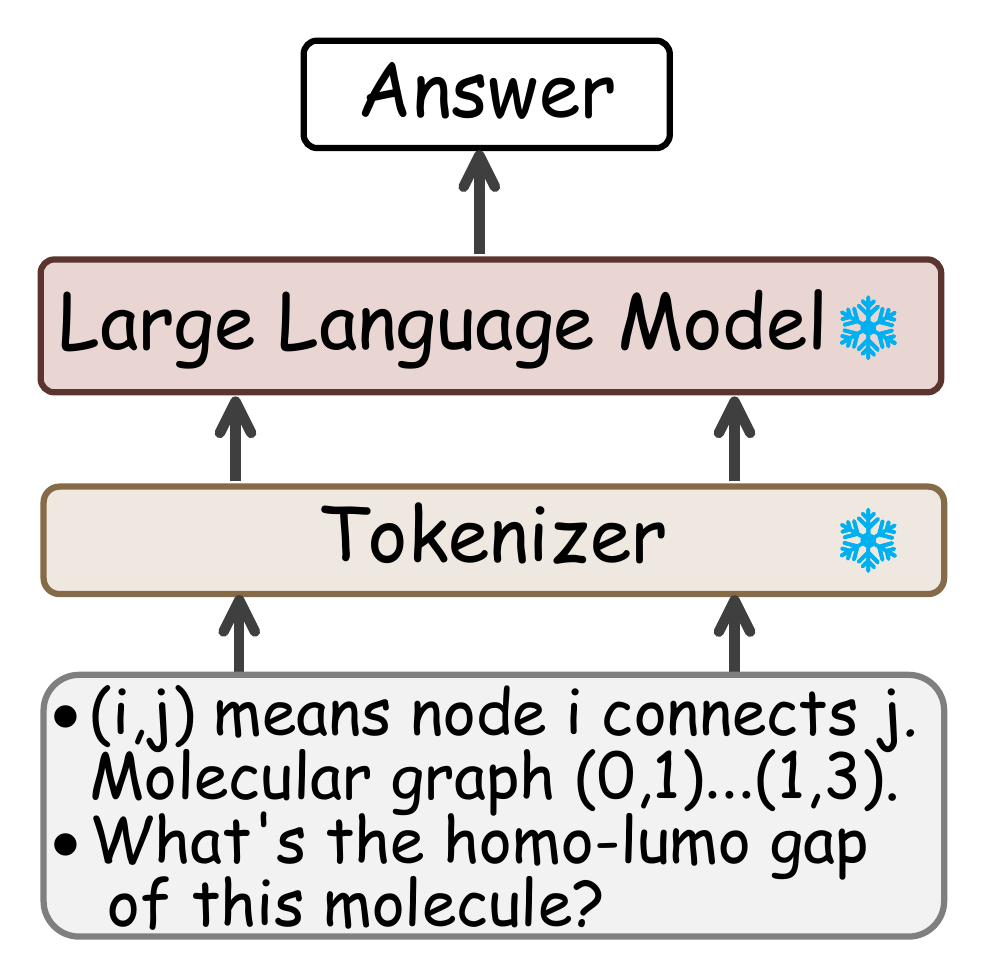}
}
\quad
\subfigure[Graph-language tuning]{
\includegraphics[width=3.8cm]{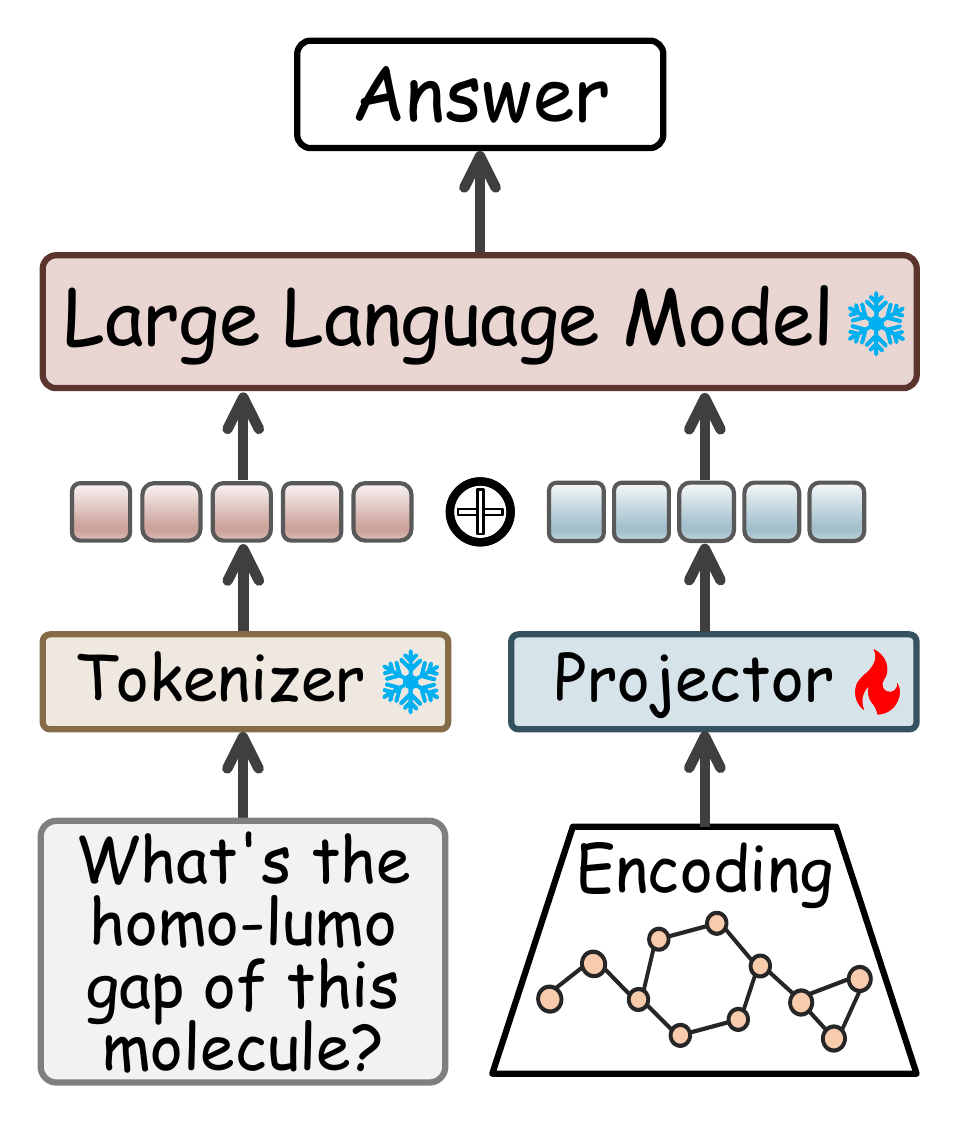}
}
\quad
\subfigure[Graph2Token (ours)]{
\includegraphics[width=3.825cm]{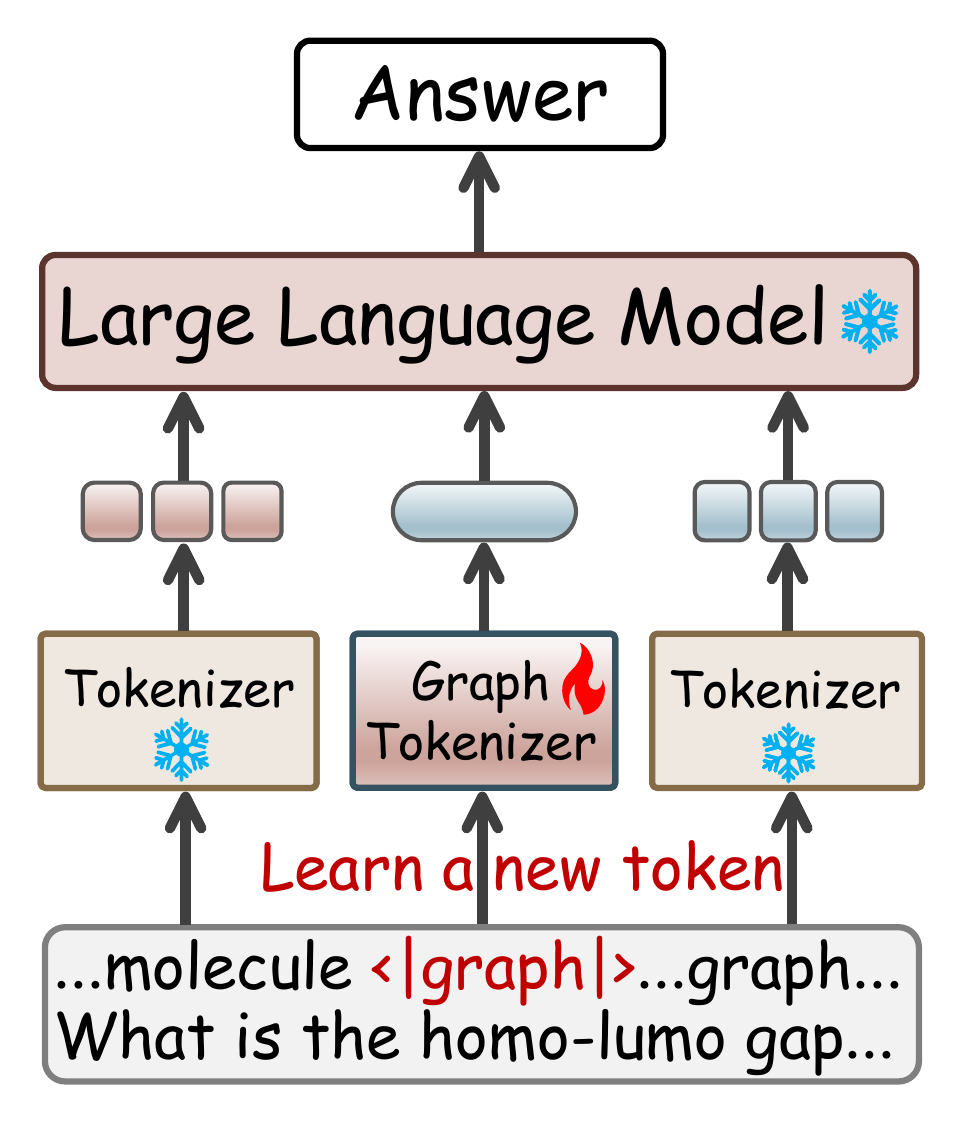}
}
\caption{Different approaches of applying LLMs to molecules.}
\label{fig:difference}
\end{figure}

To apply LLMs for molecular tasks, existing solutions often involve converting molecular structures into a format that can be processed by LLMs.
One common approach is to use the Simplified Molecular Input Line Entry System (SMILES) or SELF-referencIng Embedded Strings (SELFIES), which represent molecules as text strings (Fig. \ref{fig:difference}.(a)).
For instance, \citet{guo2023can} use SMILES as molecular representations and employ in-context learning to guide ChatGPT in understanding molecular structures.
However, a significant limitation is that LLMs often lack a proper understanding of molecular representations in SMILES strings, which in many cases leads to inaccurate or inconsistent results \cite{guo2023can}.
Therefore, parameter-efficient fine-tuning is applied to enhance LLMs' comprehension of molecular text representations \cite{fang2023mol}. 
However, this method tends to overlook molecular structure and inevitably weakens the generalization ability of LLMs by altering their semantic space during fine-tuning.
This underscores the limitations of using text data to represent molecules within the context of LLMs.

Another line of methods implicitly leverages molecular graph structure information by converting graph structure into textual representations before feeding it to the model (Fig. \ref{fig:difference}.(b)). 
These methods typically involve describing the adjacency relationships between nodes of the graph and representing the properties of nodes using text \cite{wang2024can, fatemitalk, zhao2023graphtext, liu2023evaluating}. 
Combined with zero-shot or few-shot learning techniques, as well as prompting methods, they guide LLMs in understanding complex structures.
However, relying solely on textual representation of structured data is insufficient for conducting graph reasoning using LLMs \cite{fatemitalk}.

Given the limitations of representing structured data in text, inspired by the success of multimodal large language models, researchers are exploring the use of graph-language tuning (Fig. \ref{fig:difference}.(c)). 
This involves leveraging the relationship between structured data and textual descriptions to align them in embedding space by fine-tuning a small number of parameters.
As illustrated in Fig. \ref{fig:difference}.(c), the core component is a trainable projector, which maps graph features into the text space. 
In existing work, various projectors are designed to align graph structures with text space using available molecular-text pairs \cite{molculeSTM}.
\citet{cao2023instructmol} utilize linear mapping, whereas \citet{liu2023molca} and \citet{li2024towards} implement Q-Former \cite{li2023blip}.

The graph-language tuning approaches usually adopt parameter-efficient fine-tuning, such as LoRA fine-tuning \cite{hu2021lora}.
Although only a small number of parameters are tuned, it still leads to the forgetting of knowledge in some tasks, affecting the model's generalization ability to a certain extent.
The reason lies in the inherent differences between graph-language models and vision-language models. 
For vision-language models, the success is largely due to the access to extensive, high-quality datasets.
For instance, InstructBLIP's visual encoder capitalizes on 400M image-text pairs \cite{radford2021learning}, while the training of its projector utilizes a refined vision-language dataset covering 26 datasets to ensure diversity, each featuring superior quality \cite{instructblip}. 
Conversely, the biological domain suffers from a scarcity of such data, unable to match the quantity and quality of data in the vision field.
Given these constraints, the challenging questions arise:{\it (1) Can we harness the rich prior knowledge inherent in LLMs to learn a molecular graph representation without fine-tuning the LLM backbone? 
(2) Will this approach preserve LLM's remarkable few-shot generalization, vital in biomolecular domains with limited samples? }

In this paper, we give an affirmative answer by proposing Graph2Token, a simple and effective solution, which generates a molecule graph token and aligns it to LLM tokens.
The key idea is to learn a graph token representation using the LLM token vocabulary.
In this way, a graph token can be naturally adapted by the LLM, without fine-tuning the LLM backbone.
Intuitively, for LLMs to comprehend an unseen graph token from scratch, it is analogous to a human expert who would associate a given unseen representation with existing prior knowledge, and then retrieve relevant information from its knowledge base rooted in their association.
Building upon this insight, we propose a novel alignment strategy that associates the molecular graph with LLM pre-trained token embeddings through cross multi-head attention, then retrieve useful contents from LLM token embeddings based on the computed attentions to represent the graph token.
To better generate a graph token representation, we construct a molecular-text paired dataset from multiple data sources (CHEBI and HMDB), aiming to augment the dataset with biomolecular data related to human metabolism.
Furthermore, we concurrently construct a dataset of molecular IUPAC name identifiers, incorporating them into the prompts to activate the LLMs' knowledge for target molecules.
Experiments results in few-shot scenarios, specifically those with large label distribution shifts and unseen tasks, show the competitive performance of Graph2Token. Our main contributions are as follows:
\begin{itemize}
	\item
	We introduce a novel concept of learning a new graph token for LLMs and propose a lightweight token alignment approach that can adapt a molecular graph token to LLMs without fine-tuning the LLM backbone.
	\item
	We construct a molecular-text dataset and IUPAC name dataset to reduce the gap between the graph and text modality.
	\item
	By extensive experiments in few-shot learning scenarios, our method achieves superior performance, even when encountering the unseen new tasks and greatly varied label distributions.
\end{itemize}

\section{Related Work}
\subsection{Textual Molecules for LLMs}

Recently, some studies have explored the application of LLMs in chemistry and materials science \cite{jablonka202314, jablonka2023gpt, castro2023large}, where the SMILES or SELFIES representations of molecules are taken as input to LLMs. 
\citet{guo2023can} establish a benchmark containing eight chemistry tasks that feed the SMILES strings to LLMs like GPT-4 \cite{achiam2023gpt}, Llama \cite{touvron2023llama2} and Galactica \cite{taylor2022galactica}, etc, to evaluate the capabilities of understanding and reasoning for molecules.
However, a significant limitation of LLMs is their lack of understanding of molecular representations in SMILES strings, which in many cases leads to inaccurate or inconsistent results.
Therefore, \citet{fang2023mol} employ Parameter-Efficient Fine-Tuning (PEFT) to train a molecule-oriented domain LLM using molecular related instructions and SELFIES strings.

Note that representing molecules solely through SMILES and SELFIES often neglects inherent structural information. 
Molecules can be naturally modeled as graphs \cite{xia2022systematic}. Some works have delved into translating graphs into natural language, thus enabling to directly apply LLMs for analysis and inference \cite{wang2024can, fatemitalk, liu2023evaluating, guo2023gpt4graph, zhao2023graphtext}. 
This kind of methods can be regarded as describing the graph as implicit structural information for LLMs to solve graph tasks \cite{fan2024graph}, e.g., the adjacency structure of molecular graphs is described and input into LLMs.
However, due to the limitation of input length, LLMs can only obtain local structural information, and long contexts may weaken the reasoning ability \cite{liu2024lost} and instruction following ability \cite{chen2024exploring} of LLMs.

\subsubsection{Graph-Language Tuning}

Graph-language tuning leverages graph-text pairs to map the graph modality to the text modality through training a portion of parameters.
This concept stems from the multi-modal large models, where diverse modalities such as images \cite{liu2024visual}, 3D point clouds \cite{panagopoulou2023x}, and videos \cite{huang2024vtimellm} are represented in the text space through instruction-tuned mapping functions, facilitated by large-scale and high-quality visual-instruction datasets.
Similar attempts have been made with text-attributed graphs \cite{tang2024graphgpt, zhang2024graphtranslator}, and molecules \cite{cao2023instructmol,liu2023molca, li2024towards, zhang2023moleculegpt}.
However, molecular graph-language fine-tuning methods have not demonstrated the same level of generalization capability as visual large language models, primarily due to the scarcity of molecular-text pair data in the biomolecular domain and the inherent complexity of tasks compared to the visual realm.

\section{Methods}

\begin{figure}[t]
\centering
\includegraphics[width=1\columnwidth]{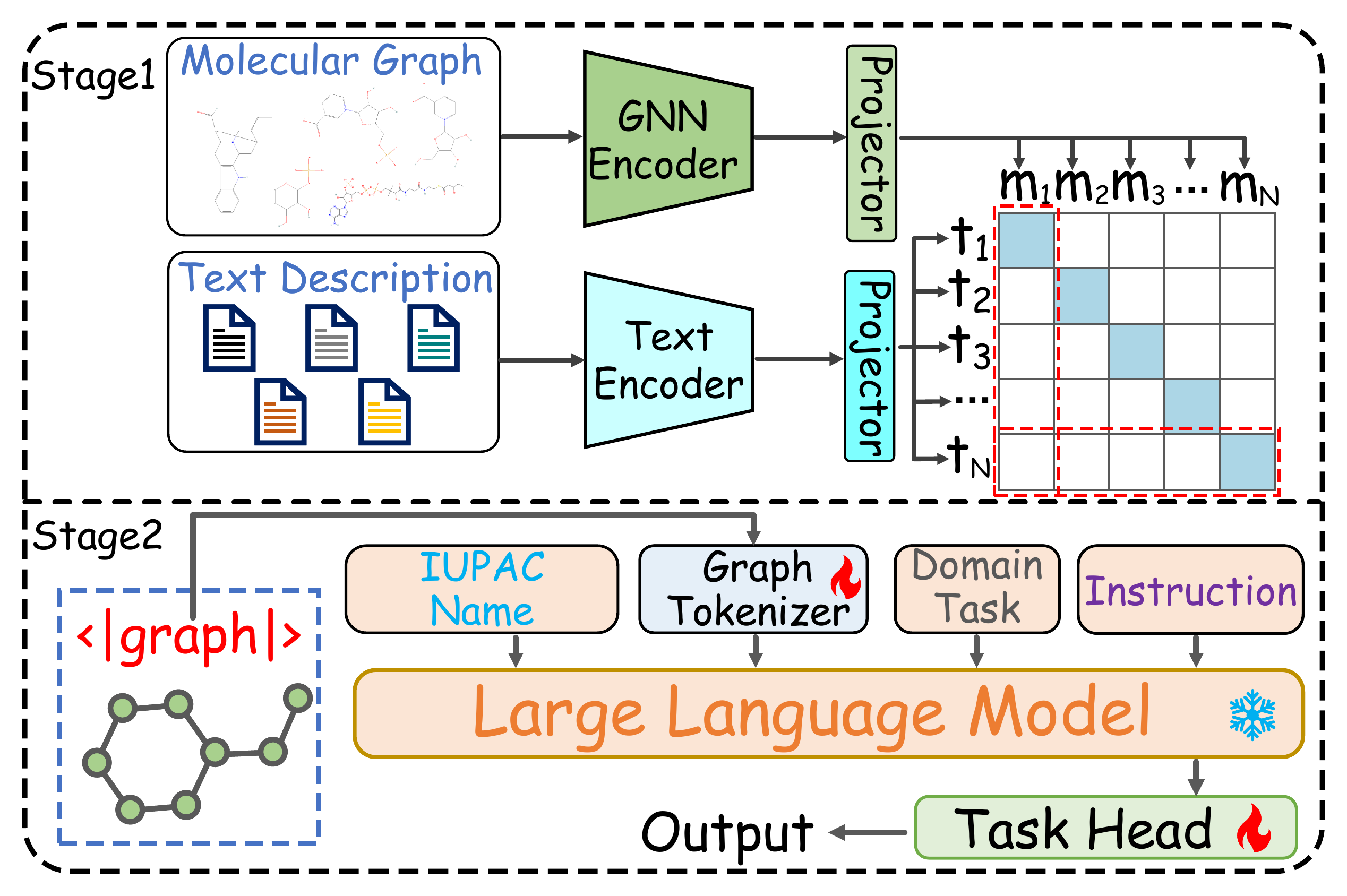} 
\caption{Overall training process of Graph2Token. Stage 1: Pre-training the molecular graph encoder based on the constructed molecular-text dataset. Stage 2: Training the graph tokenizer that can align a graph token to LLM tokens.}
\label{fig:stage}
\end{figure}

The main idea of our approach is to encode a graph into a token, and leverage the pre-trained vocabulary of LLMs to learn a new graph token representation.
In this way, it enables the model to transfer from unknown to known contexts without fine-tuning the LLM backbone.
The model involves two training stages as shown in Fig. \ref{fig:stage}: the first stage trains a molecular graph encoder for encoding graph structure and transforming textual semantics; the second stage utilizes this encoder to learn a graph tokenizer, converting unknown graph tokens into the LLM tokens.

\begin{figure*}[t]
\centering
\includegraphics[width=1.9\columnwidth]{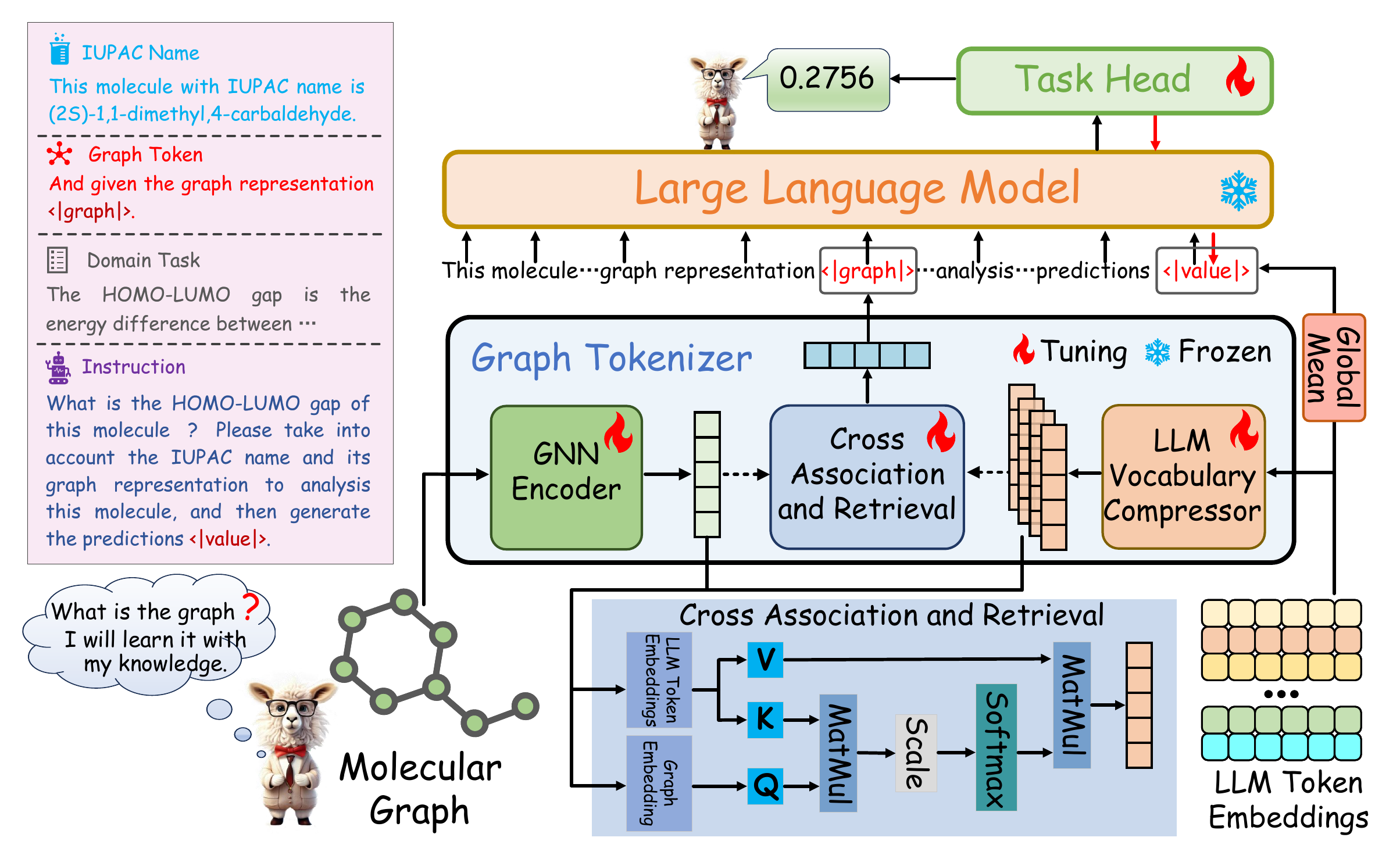}
\caption{Illustration of Graph2Token’s architecture on aligning a graph token with LLM vocabulary. Given an input molecular graph, the graph tokenizer first embeds it via pre-trained graph encoder. Then the graph features as the query state associate the compressed LLM token embeddings and retrieve the useful information according to the computed association. To activate the LLM's reasoning ability, IUPAC name and domain tasks are incorporated within the prompt. Finally, the task head outputs predicted values for specific tasks. We can see that Graph2Token doesn't fine-tune LLM backbone.}
\label{fig:align}
\end{figure*}

\subsection{Multi-source Molecular-Text Dataset and Pretrained Graph Encoder}

To train a molecular graph encoder that reduces the distance between graphs and texts representations in the feature space, we first construct a multi-source molecular-text dataset, integrating molecular-description pairs from CHEBI \cite{degtyarenko2007chebi} and HMDB \cite{wishart2022hmdb}. 
Existing molecular-description datasets typically originate from CHEBI and Pubchem \cite{wang2009pubchem}, focusing on common chemical small molecules with annotations by domain experts.
HMDB extends the scope to human metabolism-related molecules, encompassing rare and newly synthesized compounds, sourced from scientific literature.
Therefore, HMDB not only augments the data sources but also enhances the representation of biological molecules within the human metabolome category, potentially boosting the predictive performance for biomolecular properties.
Our consolidated molecular-description dataset can be used for training an efficient molecular graph encoder.

The training of stage 1 follows a similar approach as the CLIP framework \cite{radford2021learning}, which fuses two modalities through contrastive learning. 
Initially, a graph encoder and a text encoder are employed to convert molecular graph structures and textual descriptions into feature representations.
Subsequently, the linear layers are appended to project feature dimensions for graph and text features, respectively.
Following CLIP's objective optimization strategy, we also utilize the InfoNCE loss function \cite{oord2018representation}, encouraging the graph structure and text representation of the same molecule within batches to cluster together while pushing mismatched pairs apart.
Specifically, we employ the Graph Isomorphism Network (GIN) \cite{xu2018powerful} as our molecular graph encoder, renowned for its model expressivity capable of achieving the 1-WL.
For text encoding, we leverage BERT \cite{devlin2018bert}, which is predominant in embedding texts.

\subsection{Aligning Graph Tokens to LLM Token Space}
In order to enable LLMs to effectively comprehend molecular patterns, current studies fine-tune a projector and LLM backbone to align molecular graphs with texts, potentially altering semantics and reducing the capacity to follow instructions for other tasks.
Different from existing approaches, we treat a graph as a special token and design a learnable tokenizer that harnesses the prior knowledge embedded in LLMs to align graphs into representations comprehensible by the LLMs, without fine-tuning the backbone. The whole framework is shown in Fig. \ref{fig:align}.

\subsubsection{LLM input prompts}
The prompt for the LLMs, as depicted in the left of Fig. \ref{fig:align}, comprises four components: the IUPAC name in blue, the graph token in red, the domain task in gray, and the instruction in purple. 
IUPAC name is prevalent as identifiers in biochemical literature and is possibly included in training corpus of LLMs.
And it inherently contains structural information.
Therefore, we collect the IUPAC name information from PubChem database and construct the molecular datasets in IUPAC version. 
By incorporating IUPAC name into prompts, it can guide the LLMs to retrieve more relevant information from their knowledge bases and facilitate the understanding of molecules.

The graph token `\textcolor{red}{\textless$|graph|$\textgreater}' within the prompt is not initially in LLM vocabulary, and will be  processed by the specifically designed graph tokenizer.
Domain task encompasses the molecular tasks, and instruction involves both the questions and our requirements posed to LLMs.
Note that molecular property prediction often involves tasks whose output typically relies on numerical data, such as HOMO-LUMO gap. 
However, LLMs focus on syntactic relations and optimize cross-entropy loss to predict tokens, which is in contrast to the continuous value distribution required by numeric-centered regression tasks.
A straightforward solution is to make the model initially generate text-based outputs that can be converted into the desired format subsequently.
Thus, we designate a placeholder `\textit{\textcolor{red}{\textless$|value|$\textgreater}}' to store predicted value within the instruction section.
Given that the special placeholder is not included in the pre-trained LLM vocabulary, we conduct global meaning of all pre-trained token embeddings and use it to place the embedding after tokenizing the placeholder `\textit{\textcolor{red}{\textless$|value|$\textgreater}}'.
As seen in Fig. \ref{fig:align}, the prompt containing `\textit{\textcolor{red}{\textless$|graph|$\textgreater}}' and `\textit{\textcolor{red}{\textless$|value|$\textgreater}}' is directly input into LLM.
Except the graph token, which is processed by a dedicated graph tokenizer, the remaining parts are handled using the LLM tokenizer.

\subsubsection{Learn a new graph token based on LLM vocabulary}
Next, we introduce how to design a graph tokenizer that adaptively aligns molecular graph representations into the LLM token space.
Specifically, we propose to generate a graph token representation through associating pretrained token embeddings from LLM vocabulary, which can be viewed as its prior knowledge.

As illustrated in Fig. \ref{fig:align}, graph tokenizer consists of three components.
A graph neural network pretrained from stage 1 transforms molecular graphs into semantically meaningful graph features $\boldsymbol{g}\in\mathbb{R}^{e}$, which act as query patterns. 
For the LLM token embeddings $\mathbf{M}\in\mathbb{R}^{\left|\mathbf{M}\right|\times D}$, where $\left|\mathbf{M}\right|$ is the number of LLM tokens and $D$ is the dimension, it's well known that LLM vocabulary possesses an extensive searchable space, exemplified by Llama2 with 32,000 tokens and Llama3 having an even larger number of 128,000 tokens.
This poses significant challenges in computing the associations between the graph token and LLM token embeddings.
Therefore, we propose a compression module to condense the token embeddings in the entire vocabulary into a fixed set of semantic vectors.
To maximize the preservation of semantic information during compression, we employ a linear mapping function to project dense token embeddings into a condensed set of textual vectors $\mathbf{M}^{\prime}\in\mathbb{R}^{\left|\mathbf{M}^{\prime}\right|\times D}$, where $\left|\mathbf{M}^{\prime}\left|\ll\right|\mathbf{M}\right|$, 
aiming to preserve as much semantic information as possible.
Subsequently, both the graph feature $\boldsymbol{g}$ and the compressed token embeddings $\mathbf{M}^{\prime}$ are fed into the designed cross association and retrieval module, which uses cross multi-head attention network.
The graph feature $\boldsymbol{g}$ is treated as query pattern while the pretrained token embeddings are key and value patterns.
Before feeding to the cross-attention network, we first map graph patterns to a common associative space with dimension $d$ using linear transformations $\mathbf{Q}=\boldsymbol{g}\mathbf{W}_{Q}$, $\mathbf{Q}\in\mathbb{R}^{1\times d}$.
Similarly, we can obtain token embeddings $\mathbf{K}=\mathbf{M}^{\prime}\mathbf{W}_{K}$, $\mathbf{K}\in\mathbb{R}^{\left|\mathbf{M}^{\prime}\right|\times d}$ and $\mathbf{V}=\mathbf{M}^{\prime}\mathbf{W}_{V}$, $\mathbf{V}\in\mathbb{R}^{\left|\mathbf{M}^{\prime}\right|\times d}$.
$\mathbf{W}_{Q}$, $\mathbf{W}_{K}$ and $\mathbf{W}_{V}$ are trainable parameters.
Then we associate the graph pattern and LLM memory pattern via computing cross-attention matrix, and retrieve from memory according to attention matrix:
\begin{equation}
\boldsymbol{g}_{k}=\mathrm{ATT}(\mathbf{Q}_{k},\mathbf{K}_{k},\mathbf{V}_{k})=\mathrm{Soft}\max(\frac{\mathbf{Q}_{k}\mathbf{K}_{k}^{\top}}{\sqrt{d}})\mathbf{V}_{k},
\end{equation}
where $k$ represents the $k$-th attention head. 
Ultimately, a linear neural network is utilized to project the updated graph token embeddings $\boldsymbol{g}$ into the dimension $D$ of the LLM token embeddings.

Following the aforementioned association and retrieval process, we obtain molecular graph token representations that are comprehensible by the LLMs, effectively replacing the placeholder vector representation `\textit{\textcolor{red}{\textless$|graph|$\textgreater}}' within the prompt. 
These graph tokens, along with other prompt embeddings, are then jointly input into the LLM backbone for further processing.

\subsection{Output and Task Head}
Upon packaging and forwarding the instructions and graph structure embeddings through the frozen LLM backbone, we discard the prefix portion to obtain the output representation aiming to adapt the graph-level tasks.
To derive the final predictions, a task head with linear projection is employed.
Task heads cater to graph regression tasks and graph classification tasks.
The optimization objective for graph regression tasks is to minimize the mean squared error, while cross-entropy loss is used for graph classification tasks.

We can see that the trainable parameters in our Graph2Token primarily consist of the graph tokenizer module and the task
head, which are negligible compared to the parameters of LLMs. By having the original parameters of the LLMs frozen, Graph2Token better preserves their inherent semantics and functionality.

\begin{table*}[t]
\centering
\setlength{\tabcolsep}{8pt}
\begin{tabular}{l|cccccr}
\toprule

Method Type & Method & BBBP ↑ & BACE ↑ & HIV ↑ & TOX21 ↑ & Avg ↑\\
\midrule

\textit{Instruction Prompt}
 & Galactica-6.7B & 53.5 & 58.4 & 72.2 & - & 61.4 \\
 & Galactica-120B & 66.1 & 61.7 & 74.5 & - & 67.4 \\
 & Vicuna-v1.5-13B-16k [4-shot] & 49.2 & 52.7 & 50.5 & - & 50.8 \\
\specialrule{0.03em}{0.35em}{0.35em}

\textit{LoRA Fine-tuning}
 & Llama-2-7B-chat & 65.6 & 74.8 & 62.3 & - & 67.6 \\
 & Vicuna-v1.3-7B & 60.1 & 68.3 & 58.1 & - & 62.6 \\
\specialrule{0.03em}{0.35em}{0.35em}

\textit{5\% Few-shot learning}
 & \cellcolor{blue!10} \textit{Graph-Based Models} & \cellcolor{blue!10} & \cellcolor{blue!10} & \cellcolor{blue!10} & \cellcolor{blue!10} & \cellcolor{blue!10} \\
 & GPF-AttrMasking & 53.1 & 58.9 & 66.9 & 64.7 & 60.9 \\
 & GPF-GCL & 52.6 & 61.0 & 62.3 & 52.0 & 57.0 \\
 & \cellcolor{blue!10} \textit{LLM-Based Models} & \cellcolor{blue!10} & \cellcolor{blue!10} & \cellcolor{blue!10} & \cellcolor{blue!10} & \cellcolor{blue!10} \\
 & Llama3-Chat-8B (bace,bbbp,hiv) & - & - & - & 59.9 & 59.9\\
 & Llama3-Chat-8B (tox21) & 60.5 & 54.3 & 63.3 & - & 59.4\\
 & MolCA-S & 58.9 & 60.2 & 66.8 & 63.2 & 62.3\\
 & MolCA-GS & 59.1 & 61.4 & 69.5 & 64.3 & 63.6 \\
 & Graph2Token (bace,bbbp,hiv) & - & - & - & \textbf{68.7} & 68.7 \\
 & Graph2Token (tox21) & \textbf{61.0} & \textbf{63.1} & \textbf{72.3} & - & 65.5 \\
\specialrule{0.03em}{0.35em}{0.35em}

\textit{10\% Few-shot learning}
 & \cellcolor{blue!10} \textit{Graph-Based Models} & \cellcolor{blue!10} & \cellcolor{blue!10} & \cellcolor{blue!10} & \cellcolor{blue!10} & \cellcolor{blue!10} \\
 & GPF-AttrMasking & 58.8 & 62.2 & 71.3 & 66.0 & 64.6 \\
 & GPF-GCL & 56.5 & 52.1 & 49.3 & 63.6 & 55.4 \\
 & \cellcolor{blue!10} \textit{LLM-Based Models} & \cellcolor{blue!10} & \cellcolor{blue!10} & \cellcolor{blue!10} & \cellcolor{blue!10} & \cellcolor{blue!10} \\
& Llama3-Chat-8B (bace,bbbp,hiv) & - & - & - & 65.6 & 65.6\\
 & Llama3-Chat-8B (tox21) & 60.8 & 58.8 & 67.4 & - & 62.3\\
 & MolCA-S & 62.5 & 62.8 & 69.0 & 66.6 & 65.2\\
 & MolCA-GS & 63.6 & 63.9 & 72.7 & 68.5 & 67.2 \\
 & Graph2Token (bace,bbbp,hiv) & - & - & - & \textbf{72.1} & 72.1 \\
 & Graph2Token (tox21) & \textbf{65.2} & \textbf{66.0} & \textbf{74.9} & - & 68.7 \\

\bottomrule
\end{tabular}
\captionsetup{format=plain}
\caption{Few-shot learning on unseen molecular classification tasks using 5\% and 10\% training data, respectively. We report the results in ROC-AUC. \textit{Instruction Prompt}: Using input prompt as shown in Fig.\ref{fig:difference}(a). Graph2Token (tox21): Training on tox21 and transfering to other three datasets. Graph2Token (bace, bbbp, hiv): Training on the synthesized dataset of BACE, BBBP, and HIV, and transfering to tox21.
The best results are in bold.}
\label{tab:cls}
\end{table*}

\begin{table*}[ht]
\centering
\setlength{\tabcolsep}{2.7pt}
\begin{tabular}{lcccccccc}
\toprule

\multirow{2}{*}{Method} & \multicolumn{4}{c}{\textit{5\% few-shot learning}} & \multicolumn{4}{c}{\textit{10\% few-shot learning}} \\
\cmidrule(r){2-5} \cmidrule(r){6-9}
& HOMO(eV)↓ & LUMO(eV)↓ & $\Delta\epsilon$(eV)↓ & Avg(eV)↓ & HOMO(eV)↓ & LUMO(eV)↓ & $\Delta\epsilon$(eV)↓ & Avg(eV)↓ \\
\midrule

\rowcolor{blue!10}
\textit{Graph-Based Models} & & & & & & & &\\
 GPF-AttrMasking & 0.550 & 0.853 & 0.880 & 0.761 & 0.545 & 0.852 & 0.866 & 0.754\\
 GPF-GCL & 0.545 & 0.849 & 0.871 & 0.755 & 0.532 & 0.854 & 0.850 & 0.729\\
 \specialrule{0.03em}{0.35em}{0.35em}
 
 \rowcolor{blue!10}
\textit{LLM-Based Models} & & & & & & & &\\
 Mol-Instructions & 1.017 & 1.153 & 1.239 & 1.136 & 0.855 & 1.121 & 0.900 & 0.959 \\
 Vicuna-7B & 0.683 & 0.795 & 0.915 & 0.798 & 0.440 & 0.706 & 0.491 & 0.546 \\
 Llama3-Chat-8B & 0.453 & 0.532 & 0.552 & 0.512 & 0.392 & 0.471 & 0.462 & 0.442\\
 MolCA-S & 0.436 & 0.461 & 0.410 & 0.436 & 0.325 & 0.363 & 0.390 & 0.359 \\
 MolCA-GS & 0.425 & 0.453 & 0.399 & 0.426 & 0.320 & 0.343 & 0.376 & 0.346 \\
 Graph2Token (ours) & \textbf{0.407} & \textbf{0.382} & \textbf{0.386} & \textbf{0.392} & \textbf{0.292} & \textbf{0.333} & \textbf{0.346} & \textbf{0.324} \\
\bottomrule

\end{tabular}
\captionsetup{format=plain}
\caption{Few-shot learning on molecular regression tasks  using 5\% and 10\% training data, respectively.
We report the results in MAE on PubchemQC-IUPAC dataset.
The best results are in bold.}
\label{tab: reg}
\end{table*}

\section{Experiments}

In this section, we conduct extensive experiments to evaluate Graph2Token for molecular property predictions. 
We try to answer these four questions:
{\bf RQ1:} Can the model handle the unseen new molecular tasks across different datasets?
{\bf RQ2:} Can Graph2Token effectively generalize to different datasets when the labels vary greatly?
{\bf RQ3:} How does each key component of Graph2Token contribute to enhancing the model's capabilities?
{\bf RQ4:} How is the tuned number of parameters of Graph2Token compared with other approaches?

\subsection{Experimental Settings}
\subsubsection{Datasets}

For RQ1, BBBP, BACE, HIV, TOX21 datasets from MoleculeNet \cite{wu2018moleculenet} are adopted that contain 15 molecular graph classification tasks.
For RQ2, QM9 \cite{ramakrishnan2014quantum} and PubchemQC \cite{nakata2017pubchemqc} datasets are used to predict energy-related properties of HOMO, LUMO, HOMO-LUMO gap ($\Delta\epsilon$) for graph regression tasks.

For domain tasks and instructions in our prompts, we follow GIMLET \cite{zhao2023gimlet} and Mol-Instructions \cite{fang2023mol}, and refine them with GPT4 \cite{achiam2023gpt}.
All the datasets are divided into training, validation, and test sets with a ratio of 8:1:1.
In graph classification, datasets follow the setting of GIMLET according to the scaffold splitting way.
We collect the IUPAC name information from Pubchem database \cite{wang2009pubchem} and construct the IUPAC version for all the above datasets.

\subsubsection{Training Setup}

The GNN is a 5-layer GIN and the employed LLM is the Llama3-8B.
As shown in Tab. \ref{fig:align}, trainable parameters are from task head and graph tokenizer including GNN, cross association and retrieval, LLM compressor.
When transferring to few-shot scenarios, we freeze the LLM compressor and only tune the remaining three parts.

\subsubsection{Baselines}

We incorporate both graph-based models and LLM-based models as baselines.
The graph-based models include graph prompt learning method GPF \cite{fang2024universal} with AttrMasking \cite{hu2019strategies} and GCL \cite{you2020graphcl} as pretrained GNN models.
For the LLM-based models, we consider the following categories: directly input the textual molecules as well as instruction prompts including
Galactica-6.7B \cite{taylor2022galactica}, Galactica-120B \cite{taylor2022galactica} and Vicuna-v1.5-13B-16K \cite{chiang2023vicuna};
LoRA fine-tuning methods on the entire training set based on Llama2-Chat-7B \cite{touvron2023llama2}, Vicuna-v1.3-7B \cite{chiang2023vicuna};
LoRA fine-tuning methods in few-shot scenarios including Llama3-Chat-8B \cite{dubey2024llama}, Vicuna-v1.5-7B \cite{chiang2023vicuna} and molecule-oriented approach Mol-Instructions \cite{fang2023mol};
graph-language tuning methods including MolCA \cite{liu2023molca} and its variant.

\subsection{Few-Shot Performance on Classification Task}

\subsubsection{Setups}

To answer RQ1, we design two experiments on molecular classification datasets across 15 tasks. 
The first experiment involves training the second stage using a synthesized dataset that includes three tasks across BACE, BBBP and HIV datasets, then evaluating the model's generalization ability on the TOX21 with 12 tasks.
The second experiment focuses on training the second stage exclusively with the TOX21 dataset, followed by evaluating the model on the BACE, BBBP, and HIV datasets individually.

\subsubsection{Results}

The results of 5\% and 10\% few-shot learning are shown in Tab. \ref{tab:cls}.
We can see that Graph2Token exceeds all baselines when confronted with the new molecular tasks.
We attribute this to the use of LLMs and coupled with a successful way of integrating graph structural information.
Indeed, the graph-language tuning model MolCA has also demonstrated impressive results, further underscoring the potential of fusing graph structures into LLMs for tackling molecular tasks.
When using 10\% of the training samples, Graph2Token achieves an average 3.7\% improvement in ROC-AUC compared to the best results of MolCA.
Remarkably, despite the limited amounts of training data, Graph2Token can match the performance of LLMs with LoRA fine-tuned on the entire training set, as seen in LoRA Fine-tuneing part in Tab. \ref{tab:cls}.
Similar trends can be observed in the 5\% few-shot learning scenarios, where our average improvement over MolCA exceeds 4\%.

\subsection{Few-Shot Performance on Regression Task}

\subsubsection{Setups}

To answer RQ2, we consolidate three tasks from the QM9 dataset—HOMO, LUMO, and $\Delta\epsilon$—into a multi-task dataset to conduct the training of stage 2.
We evaluate on few-shot scenarios with a subset of training data (5\% and 10\% molecular samples) on the PubchemQC-IUPAC dataset.
For both Vicuna-7B and Llama3-Chat-8B, we employ SMILES as the molecular representation and fine-tune their backbones with LoRA on the consolidated QM9 dataset, and then transfer to the PubchemQC-IUPAC.
Mol-Instructions has been trained on QM9, and therefore we directly transfer it to PubchemQC-IUPAC.

\subsubsection{Results}

The results of 5\% and 10\% few-shot scenarios are presented in Tab. \ref{tab: reg}.
We can see that Graph2Token outperforms all baselines, which underscores the effectiveness of our strategy in aligning a graph token into the manner that LLM can understand.
Graph2Token realizes average 7.1\% and 6.4\% reductions in comparison to graph-language tuning method MolCA.
Compared to Llama3-Chat with LoRA fine-tuning, our average enhancements are 20\% and 16\%. 
We also observe that, in the case where labels vary greatly, LLMs frequently exhibit hallucinations, generating responses that appear plausible yet deviate from factfulness. 
Graph2Token can adapt to new datasets with limited labeled molecular samples, demonstrating remarkable few-shot generalization capabilities.

\begin{table}[h]
\centering
\setlength{\tabcolsep}{7pt}
\begin{tabular}{lccc}
\toprule

Method & BBBP ↑ & TOX21 ↑ & Avg ↑\\
\midrule

\rowcolor{blue!10}
\textit{5\% few-shot learning} & & &\\
 w/o IUPAC & 59.2 & 67.8 & 63.5\\
 w/o Alignment & 58.5 & 66.7 & 62.6\\
 w/o MT Data & 60.4 & 66.8 & 63.6\\
 w/o LLM & 56.9 & 65.5 & 61.2\\
 Graph2Token(ours) & \textbf{61.0} & \textbf{68.7} & \textbf{64.9}\\
\specialrule{0.03em}{0.35em}{0.35em}

\rowcolor{blue!10}
\textit{10\% few-shot learning} & & &\\
 w/o IUPAC & 63.5 & 71.0 & 67.3\\
 w/o Alignment & 62.1 & 69.4 & 65.8\\
 w/o MT Data & 63.3 & 70.3 & 66.8\\
 w/o LLM & 60.0 & 66.7 & 63.4\\
 Graph2Token(ours) & \textbf{65.2} & \textbf{72.1} & \textbf{68.7}\\
\bottomrule

\end{tabular}
\captionsetup{format=plain}
\caption{Ablation results (ROC-AUC) on molecular classification tasks on BBBP and TOX21 datasets. }
\label{tab: ablation}
\end{table}

\begin{table}[h]
\centering
\setlength{\tabcolsep}{6pt}
\begin{tabular}{l|c|r}
\toprule

Method & Param. (M) & \makecell[c]{Train Ratio. (\%)} \\
\midrule

Graph2Token & 8.60 & \makecell[c]{0.11} \\
MolCA & 109.09 & \makecell[c]{7.65} \\
Mol-Instructions & 39.98 & \makecell[c]{0.59} \\
Llama3-Chat-8B & 88.89 & \makecell[c]{1.03} \\
Vicuna-7B & 79.95 & \makecell[c]{1.17} \\
\bottomrule

\end{tabular}
\captionsetup{format=plain}
\caption{Number of tuned parameters of Graph2Token compared with LLM-Based Models.}
\label{tab: parameter}
\end{table}

\subsection{Ablation Study}

This section addresses RQ3 by investigating how the key components of Graph2Token contribute to the performance, i.e., IUPAC name in prompt (-\textbf{IUPAC}), cross association and retrieval for alignment (-\textbf{Alignment}), Molecular-Text data for pretrained GNN (-\textbf{MT Data}) and LLM (-\textbf{LLM}).

As shown in Tab. \ref{tab: ablation}, when removing the LLM and only retaining the pretrained GNN with a linear layer for specific tasks, there is a sharp performance decline, indicating that our strategy, adding a graph tokenizer without fine-tuning the LLM backbone, can maintain the remarkable few-shot generalization ability of LLMs.
For the alignment, we replace the cross-association and retrieval with a single linear mapping to maintain the feature dimensions transformation.
The results show that the LLM vocabulary can serve as prior knowledge, with cross attention calculating the relevance to the unknown graph token.
In this way, useful embedding information is retrieved based on the calculated association that can be leveraged to represent graph features in a LLMs comprehensible manner.
By replacing the graph encoder pre-trained on MT Data with random initialization, we observe a performance drop, indicating the importance of bridging the feature gap between graph and text modality.
Furthermore, by incorporating IUPAC name, we activate the LLMs' relevant knowledge of target molecules, which leads to improved performance.

\subsection{Efficiency Study}

For RQ 4, Tab. \ref{tab: parameter} shows the number of tuned parameters of Graph2Token and other LLM-Based baselines.
This comparison reveals that our graph alignment approach serves as a lightweight solution, empowering LLMs to comprehend graph structures effectively.
As mentioned in the training setup, under the few-shot scenarios, we only tune the graph encoder, cross association and retrieval along with task head, where the trainable parameters account for merely 0.11\% of the entire framework.
While LoRA fine-tuning represents a parameter-efficient approach to adapt the LLM backbone to vertical domains, our method still prevails in balancing performance and efficiency.

\section{Conclusion}
In this paper, we propose Graph2Token, which bridges molecular graphs and LLMs via aligning a graph token to LLM token space. 
Graph2Token is a lightweight solution without fine-tuning LLM backbone.
To achieve this, we first construct a molecular-text dataset from multiple resources, as well as a molecular IUPAC name dataset.
Then, we design a cross association and retrieval module to align the graph representations with pre-trained LLM token embeddings.
Evaluations on few-shot learning scenarios demonstrate Graph2Token maintains the few-shot generalization ability of LLMs and effectively solves the data scarcity issue in the molecular field.

\section*{Acknowledgements}

This work was supported by the National Natural Science Foundation of China under Grant 62276044.


\bibliography{aaai25}

\begin{thebibliography}{47}
\providecommand{\natexlab}[1]{#1}

\bibitem[{Achiam et~al.(2023)Achiam, Adler, Agarwal, Ahmad, Akkaya, Aleman, Almeida, Altenschmidt, Altman, Anadkat et~al.}]{achiam2023gpt}
Achiam, J.; Adler, S.; Agarwal, S.; Ahmad, L.; Akkaya, I.; Aleman, F.~L.; Almeida, D.; Altenschmidt, J.; Altman, S.; Anadkat, S.; et~al. 2023.
\newblock Gpt-4 technical report.
\newblock \emph{arXiv preprint arXiv:2303.08774}.

\bibitem[{Cao et~al.(2023)Cao, Liu, Lu, Yao, and Li}]{cao2023instructmol}
Cao, H.; Liu, Z.; Lu, X.; Yao, Y.; and Li, Y. 2023.
\newblock Instructmol: Multi-modal integration for building a versatile and reliable molecular assistant in drug discovery.
\newblock \emph{arXiv preprint arXiv:2311.16208}.

\bibitem[{Castro~Nascimento and Pimentel(2023)}]{castro2023large}
Castro~Nascimento, C.~M.; and Pimentel, A.~S. 2023.
\newblock Do large language models understand chemistry? a conversation with chatgpt.
\newblock \emph{Journal of Chemical Information and Modeling}, 63(6): 1649--1655.

\bibitem[{Chen et~al.(2024{\natexlab{a}})Chen, Zhao, JAISWAL, Shah, and Wang}]{chenllaga}
Chen, R.; Zhao, T.; JAISWAL, A.~K.; Shah, N.; and Wang, Z. 2024{\natexlab{a}}.
\newblock LLaGA: Large Language and Graph Assistant.
\newblock In \emph{Forty-first International Conference on Machine Learning}.

\bibitem[{Chen et~al.(2024{\natexlab{b}})Chen, Mao, Li, Jin, Wen, Wei, Wang, Yin, Fan, Liu et~al.}]{chen2024exploring}
Chen, Z.; Mao, H.; Li, H.; Jin, W.; Wen, H.; Wei, X.; Wang, S.; Yin, D.; Fan, W.; Liu, H.; et~al. 2024{\natexlab{b}}.
\newblock Exploring the potential of large language models (llms) in learning on graphs.
\newblock \emph{ACM SIGKDD Explorations Newsletter}, 25(2): 42--61.

\bibitem[{Chiang et~al.(2023)Chiang, Li, Lin, Sheng, Wu, Zhang, Zheng, Zhuang, Zhuang, Gonzalez et~al.}]{chiang2023vicuna}
Chiang, W.-L.; Li, Z.; Lin, Z.; Sheng, Y.; Wu, Z.; Zhang, H.; Zheng, L.; Zhuang, S.; Zhuang, Y.; Gonzalez, J.~E.; et~al. 2023.
\newblock Vicuna: An open-source chatbot impressing gpt-4 with 90\%* chatgpt quality.
\newblock \emph{See https://vicuna. lmsys. org (accessed 14 April 2023)}, 2(3): 6.

\bibitem[{Dai et~al.(2023)Dai, Li, Li, Tiong, Zhao, Wang, Li, Fung, and Hoi}]{instructblip}
Dai, W.; Li, J.; Li, D.; Tiong, A. M.~H.; Zhao, J.; Wang, W.; Li, B.; Fung, P.; and Hoi, S. 2023.
\newblock InstructBLIP: Towards General-purpose Vision-Language Models with Instruction Tuning.
\newblock arXiv:2305.06500.

\bibitem[{Degtyarenko et~al.(2007)Degtyarenko, De~Matos, Ennis, Hastings, Zbinden, McNaught, Alc{\'a}ntara, Darsow, Guedj, and Ashburner}]{degtyarenko2007chebi}
Degtyarenko, K.; De~Matos, P.; Ennis, M.; Hastings, J.; Zbinden, M.; McNaught, A.; Alc{\'a}ntara, R.; Darsow, M.; Guedj, M.; and Ashburner, M. 2007.
\newblock ChEBI: a database and ontology for chemical entities of biological interest.
\newblock \emph{Nucleic acids research}, 36(suppl\_1): D344--D350.

\bibitem[{Devlin et~al.(2018)Devlin, Chang, Lee, and Toutanova}]{devlin2018bert}
Devlin, J.; Chang, M.-W.; Lee, K.; and Toutanova, K. 2018.
\newblock Bert: Pre-training of deep bidirectional transformers for language understanding.
\newblock \emph{arXiv preprint arXiv:1810.04805}.

\bibitem[{Dubey et~al.(2024)Dubey, Jauhri, Pandey, Kadian, Al-Dahle, Letman, Mathur, Schelten, Yang, Fan et~al.}]{dubey2024llama}
Dubey, A.; Jauhri, A.; Pandey, A.; Kadian, A.; Al-Dahle, A.; Letman, A.; Mathur, A.; Schelten, A.; Yang, A.; Fan, A.; et~al. 2024.
\newblock The llama 3 herd of models.
\newblock \emph{arXiv preprint arXiv:2407.21783}.

\bibitem[{Fan et~al.(2024)Fan, Wang, Huang, Chen, Song, Tang, Mao, Liu, Liu, Yin et~al.}]{fan2024graph}
Fan, W.; Wang, S.; Huang, J.; Chen, Z.; Song, Y.; Tang, W.; Mao, H.; Liu, H.; Liu, X.; Yin, D.; et~al. 2024.
\newblock Graph machine learning in the era of large language models (llms).
\newblock \emph{arXiv preprint arXiv:2404.14928}.

\bibitem[{Fang et~al.(2024)Fang, Zhang, Yang, Wang, and Chen}]{fang2024universal}
Fang, T.; Zhang, Y.; Yang, Y.; Wang, C.; and Chen, L. 2024.
\newblock Universal prompt tuning for graph neural networks.
\newblock \emph{Advances in Neural Information Processing Systems}, 36.

\bibitem[{Fang et~al.(2023)Fang, Liang, Zhang, Liu, Huang, Chen, Fan, and Chen}]{fang2023mol}
Fang, Y.; Liang, X.; Zhang, N.; Liu, K.; Huang, R.; Chen, Z.; Fan, X.; and Chen, H. 2023.
\newblock Mol-instructions: A large-scale biomolecular instruction dataset for large language models.
\newblock \emph{arXiv preprint arXiv:2306.08018}.

\bibitem[{Fatemi, Halcrow, and Perozzi(2024)}]{fatemitalk}
Fatemi, B.; Halcrow, J.; and Perozzi, B. 2024.
\newblock Talk like a Graph: Encoding Graphs for Large Language Models.
\newblock \emph{International Conference on Learning Representations}.

\bibitem[{Guo et~al.(2023{\natexlab{a}})Guo, Du, Liu, Zhou, He, and Han}]{guo2023gpt4graph}
Guo, J.; Du, L.; Liu, H.; Zhou, M.; He, X.; and Han, S. 2023{\natexlab{a}}.
\newblock Gpt4graph: Can large language models understand graph structured data? an empirical evaluation and benchmarking.
\newblock \emph{arXiv preprint arXiv:2305.15066}.

\bibitem[{Guo et~al.(2023{\natexlab{b}})Guo, Nan, Liang, Guo, Chawla, Wiest, Zhang et~al.}]{guo2023can}
Guo, T.; Nan, B.; Liang, Z.; Guo, Z.; Chawla, N.; Wiest, O.; Zhang, X.; et~al. 2023{\natexlab{b}}.
\newblock What can large language models do in chemistry? a comprehensive benchmark on eight tasks.
\newblock \emph{Advances in Neural Information Processing Systems}, 36: 59662--59688.

\bibitem[{Hu et~al.(2021)Hu, Shen, Wallis, Allen-Zhu, Li, Wang, Wang, and Chen}]{hu2021lora}
Hu, E.~J.; Shen, Y.; Wallis, P.; Allen-Zhu, Z.; Li, Y.; Wang, S.; Wang, L.; and Chen, W. 2021.
\newblock Lora: Low-rank adaptation of large language models.
\newblock \emph{arXiv preprint arXiv:2106.09685}.

\bibitem[{Hu et~al.(2019)Hu, Liu, Gomes, Zitnik, Liang, Pande, and Leskovec}]{hu2019strategies}
Hu, W.; Liu, B.; Gomes, J.; Zitnik, M.; Liang, P.; Pande, V.; and Leskovec, J. 2019.
\newblock Strategies for pre-training graph neural networks.
\newblock \emph{arXiv preprint arXiv:1905.12265}.

\bibitem[{Huang et~al.(2024)Huang, Wang, Chen, Song, and Zhu}]{huang2024vtimellm}
Huang, B.; Wang, X.; Chen, H.; Song, Z.; and Zhu, W. 2024.
\newblock Vtimellm: Empower llm to grasp video moments.
\newblock In \emph{Proceedings of the IEEE/CVF Conference on Computer Vision and Pattern Recognition}, 14271--14280.

\bibitem[{Jablonka et~al.(2023{\natexlab{a}})Jablonka, Ai, Al-Feghali, Badhwar, Bocarsly, Bran, Bringuier, Brinson, Choudhary, Circi et~al.}]{jablonka202314}
Jablonka, K.~M.; Ai, Q.; Al-Feghali, A.; Badhwar, S.; Bocarsly, J.~D.; Bran, A.~M.; Bringuier, S.; Brinson, L.~C.; Choudhary, K.; Circi, D.; et~al. 2023{\natexlab{a}}.
\newblock 14 examples of how LLMs can transform materials science and chemistry: a reflection on a large language model hackathon.
\newblock \emph{Digital Discovery}, 2(5): 1233--1250.

\bibitem[{Jablonka et~al.(2023{\natexlab{b}})Jablonka, Schwaller, Ortega-Guerrero, and Smit}]{jablonka2023gpt}
Jablonka, K.~M.; Schwaller, P.; Ortega-Guerrero, A.; and Smit, B. 2023{\natexlab{b}}.
\newblock Is GPT-3 all you need for low-data discovery in chemistry?

\bibitem[{Li et~al.(2023)Li, Li, Savarese, and Hoi}]{li2023blip}
Li, J.; Li, D.; Savarese, S.; and Hoi, S. 2023.
\newblock Blip-2: Bootstrapping language-image pre-training with frozen image encoders and large language models.
\newblock In \emph{International conference on machine learning}, 19730--19742. PMLR.

\bibitem[{Li et~al.(2024)Li, Liu, Luo, Wang, He, Kawaguchi, Chua, and Tian}]{li2024towards}
Li, S.; Liu, Z.; Luo, Y.; Wang, X.; He, X.; Kawaguchi, K.; Chua, T.-S.; and Tian, Q. 2024.
\newblock Towards 3d molecule-text interpretation in language models.
\newblock \emph{International Conference on Learning Representations}.

\bibitem[{Liu and Wu(2023)}]{liu2023evaluating}
Liu, C.; and Wu, B. 2023.
\newblock Evaluating large language models on graphs: Performance insights and comparative analysis.
\newblock \emph{arXiv preprint arXiv:2308.11224}.

\bibitem[{Liu et~al.(2024{\natexlab{a}})Liu, Li, Wu, and Lee}]{liu2024visual}
Liu, H.; Li, C.; Wu, Q.; and Lee, Y.~J. 2024{\natexlab{a}}.
\newblock Visual instruction tuning.
\newblock \emph{Advances in neural information processing systems}, 36.

\bibitem[{Liu et~al.(2024{\natexlab{b}})Liu, Lin, Hewitt, Paranjape, Bevilacqua, Petroni, and Liang}]{liu2024lost}
Liu, N.~F.; Lin, K.; Hewitt, J.; Paranjape, A.; Bevilacqua, M.; Petroni, F.; and Liang, P. 2024{\natexlab{b}}.
\newblock Lost in the middle: How language models use long contexts.
\newblock \emph{Transactions of the Association for Computational Linguistics}, 12: 157--173.

\bibitem[{Liu et~al.(2023{\natexlab{a}})Liu, Nie, Wang, Lu, Qiao, Liu, Tang, Xiao, and Anandkumar}]{molculeSTM}
Liu, S.; Nie, W.; Wang, C.; Lu, J.; Qiao, Z.; Liu, L.; Tang, J.; Xiao, C.; and Anandkumar, A. 2023{\natexlab{a}}.
\newblock Multi-modal molecule structure--text model for text-based retrieval and editing.
\newblock \emph{Nature Machine Intelligence}, 5(12): 1447--1457.

\bibitem[{Liu et~al.(2023{\natexlab{b}})Liu, Li, Luo, Fei, Cao, Kawaguchi, Wang, and Chua}]{liu2023molca}
Liu, Z.; Li, S.; Luo, Y.; Fei, H.; Cao, Y.; Kawaguchi, K.; Wang, X.; and Chua, T.-S. 2023{\natexlab{b}}.
\newblock Molca: Molecular graph-language modeling with cross-modal projector and uni-modal adapter.
\newblock \emph{arXiv preprint arXiv:2310.12798}.

\bibitem[{Nakata and Shimazaki(2017)}]{nakata2017pubchemqc}
Nakata, M.; and Shimazaki, T. 2017.
\newblock PubChemQC project: a large-scale first-principles electronic structure database for data-driven chemistry.
\newblock \emph{Journal of chemical information and modeling}, 57(6): 1300--1308.

\bibitem[{Oord, Li, and Vinyals(2018)}]{oord2018representation}
Oord, A. v.~d.; Li, Y.; and Vinyals, O. 2018.
\newblock Representation learning with contrastive predictive coding.
\newblock \emph{arXiv preprint arXiv:1807.03748}.

\bibitem[{Panagopoulou et~al.(2023)Panagopoulou, Xue, Yu, Li, Li, Joty, Xu, Savarese, Xiong, and Niebles}]{panagopoulou2023x}
Panagopoulou, A.; Xue, L.; Yu, N.; Li, J.; Li, D.; Joty, S.; Xu, R.; Savarese, S.; Xiong, C.; and Niebles, J.~C. 2023.
\newblock X-instructblip: A framework for aligning x-modal instruction-aware representations to llms and emergent cross-modal reasoning.
\newblock \emph{arXiv preprint arXiv:2311.18799}.

\bibitem[{Radford et~al.(2021)Radford, Kim, Hallacy, Ramesh, Goh, Agarwal, Sastry, Askell, Mishkin, Clark et~al.}]{radford2021learning}
Radford, A.; Kim, J.~W.; Hallacy, C.; Ramesh, A.; Goh, G.; Agarwal, S.; Sastry, G.; Askell, A.; Mishkin, P.; Clark, J.; et~al. 2021.
\newblock Learning transferable visual models from natural language supervision.
\newblock In \emph{International conference on machine learning}, 8748--8763. PMLR.

\bibitem[{Ramakrishnan et~al.(2014)Ramakrishnan, Dral, Rupp, and Von~Lilienfeld}]{ramakrishnan2014quantum}
Ramakrishnan, R.; Dral, P.~O.; Rupp, M.; and Von~Lilienfeld, O.~A. 2014.
\newblock Quantum chemistry structures and properties of 134 kilo molecules.
\newblock \emph{Scientific data}, 1(1): 1--7.

\bibitem[{Tang et~al.(2024)Tang, Yang, Wei, Shi, Su, Cheng, Yin, and Huang}]{tang2024graphgpt}
Tang, J.; Yang, Y.; Wei, W.; Shi, L.; Su, L.; Cheng, S.; Yin, D.; and Huang, C. 2024.
\newblock Graphgpt: Graph instruction tuning for large language models.
\newblock In \emph{Proceedings of the 47th International ACM SIGIR Conference on Research and Development in Information Retrieval}, 491--500.

\bibitem[{Taylor et~al.(2022)Taylor, Kardas, Cucurull, Scialom, Hartshorn, Saravia, Poulton, Kerkez, and Stojnic}]{taylor2022galactica}
Taylor, R.; Kardas, M.; Cucurull, G.; Scialom, T.; Hartshorn, A.; Saravia, E.; Poulton, A.; Kerkez, V.; and Stojnic, R. 2022.
\newblock Galactica: A large language model for science.
\newblock \emph{arXiv preprint arXiv:2211.09085}.

\bibitem[{Touvron et~al.(2023)Touvron, Martin, Stone, Albert, Almahairi, Babaei, Bashlykov, Batra, Bhargava, Bhosale et~al.}]{touvron2023llama2}
Touvron, H.; Martin, L.; Stone, K.; Albert, P.; Almahairi, A.; Babaei, Y.; Bashlykov, N.; Batra, S.; Bhargava, P.; Bhosale, S.; et~al. 2023.
\newblock Llama 2: Open foundation and fine-tuned chat models.
\newblock \emph{arXiv preprint arXiv:2307.09288}.

\bibitem[{Wang et~al.(2024)Wang, Feng, He, Tan, Han, and Tsvetkov}]{wang2024can}
Wang, H.; Feng, S.; He, T.; Tan, Z.; Han, X.; and Tsvetkov, Y. 2024.
\newblock Can language models solve graph problems in natural language?
\newblock \emph{Advances in Neural Information Processing Systems}, 36.

\bibitem[{Wang et~al.(2009)Wang, Xiao, Suzek, Zhang, Wang, and Bryant}]{wang2009pubchem}
Wang, Y.; Xiao, J.; Suzek, T.~O.; Zhang, J.; Wang, J.; and Bryant, S.~H. 2009.
\newblock PubChem: a public information system for analyzing bioactivities of small molecules.
\newblock \emph{Nucleic acids research}, 37(suppl\_2): W623--W633.

\bibitem[{Wishart et~al.(2022)Wishart, Guo, Oler, Wang, Anjum, Peters, Dizon, Sayeeda, Tian, Lee et~al.}]{wishart2022hmdb}
Wishart, D.~S.; Guo, A.; Oler, E.; Wang, F.; Anjum, A.; Peters, H.; Dizon, R.; Sayeeda, Z.; Tian, S.; Lee, B.~L.; et~al. 2022.
\newblock HMDB 5.0: the human metabolome database for 2022.
\newblock \emph{Nucleic acids research}, 50(D1): D622--D631.

\bibitem[{Wu et~al.(2018)Wu, Ramsundar, Feinberg, Gomes, Geniesse, Pappu, Leswing, and Pande}]{wu2018moleculenet}
Wu, Z.; Ramsundar, B.; Feinberg, E.~N.; Gomes, J.; Geniesse, C.; Pappu, A.~S.; Leswing, K.; and Pande, V. 2018.
\newblock MoleculeNet: a benchmark for molecular machine learning.
\newblock \emph{Chemical science}, 9(2): 513--530.

\bibitem[{Xia et~al.(2022)Xia, Zhu, Du, Liu, and Li}]{xia2022systematic}
Xia, J.; Zhu, Y.; Du, Y.; Liu, Y.; and Li, S.~Z. 2022.
\newblock A systematic survey of molecular pre-trained models.
\newblock \emph{arXiv preprint arXiv:2210.16484}.

\bibitem[{Xu et~al.(2018)Xu, Hu, Leskovec, and Jegelka}]{xu2018powerful}
Xu, K.; Hu, W.; Leskovec, J.; and Jegelka, S. 2018.
\newblock How powerful are graph neural networks?
\newblock \emph{arXiv preprint arXiv:1810.00826}.

\bibitem[{You et~al.(2020)You, Chen, Sui, Chen, Wang, and Shen}]{you2020graphcl}
You, Y.; Chen, T.; Sui, Y.; Chen, T.; Wang, Z.; and Shen, Y. 2020.
\newblock Graph contrastive learning with augmentations.
\newblock \emph{Advances in neural information processing systems}, 33: 5812--5823.

\bibitem[{Zhang et~al.(2024)Zhang, Sun, Wang, Fan, Mo, Xu, Liu, Yang, and Shi}]{zhang2024graphtranslator}
Zhang, M.; Sun, M.; Wang, P.; Fan, S.; Mo, Y.; Xu, X.; Liu, H.; Yang, C.; and Shi, C. 2024.
\newblock GraphTranslator: Aligning Graph Model to Large Language Model for Open-ended Tasks.
\newblock In \emph{Proceedings of the ACM on Web Conference 2024}, 1003--1014.

\bibitem[{Zhang et~al.(2023)Zhang, Wang, Nie, Eaton, Rees, and Gu}]{zhang2023moleculegpt}
Zhang, W.; Wang, X.; Nie, W.; Eaton, J.; Rees, B.; and Gu, Q. 2023.
\newblock MoleculeGPT: Instruction following large language models for molecular property prediction.
\newblock In \emph{NeurIPS 2023 Workshop on New Frontiers of AI for Drug Discovery and Development}.

\bibitem[{Zhao et~al.(2023{\natexlab{a}})Zhao, Liu, Chang, Xu, Fu, Deng, Kong, and Liu}]{zhao2023gimlet}
Zhao, H.; Liu, S.; Chang, M.; Xu, H.; Fu, J.; Deng, Z.; Kong, L.; and Liu, Q. 2023{\natexlab{a}}.
\newblock Gimlet: A unified graph-text model for instruction-based molecule zero-shot learning.
\newblock \emph{Advances in Neural Information Processing Systems}, 36: 5850--5887.

\bibitem[{Zhao et~al.(2023{\natexlab{b}})Zhao, Zhuo, Shen, Qu, Liu, Bronstein, Zhu, and Tang}]{zhao2023graphtext}
Zhao, J.; Zhuo, L.; Shen, Y.; Qu, M.; Liu, K.; Bronstein, M.; Zhu, Z.; and Tang, J. 2023{\natexlab{b}}.
\newblock Graphtext: Graph reasoning in text space.
\newblock \emph{arXiv preprint arXiv:2310.01089}.

\end{thebibliography}

\newpage

\end{document}